\documentclass{article}

\usepackage{times}
\usepackage{graphicx} 
\usepackage{subfigure} 

\usepackage{natbib}

\usepackage{algorithm}
\usepackage{algorithmic}

\usepackage{hyperref}

\usepackage{tikz}
\usetikzlibrary{decorations.pathreplacing,calc,shapes,positioning}
\tikzset{>=latex}

\usepackage[accepted]{icml2016}
\usepackage{balance}  
\usepackage{amsmath}
\usepackage{amsfonts}
\usepackage{bm}

\def\1{\bm{1}}

\def\rvx{{\mathbf{x}}}

\def\rvz{{\mathbf{z}}}

\def\vzero{{\bm{0}}}

\def\vd{{\bm{d}}}

\def\vf{{\bm{f}}}
\def\vg{{\bm{g}}}

\def\vx{{\bm{x}}}

\def\vz{{\bm{z}}}

\def\vmu{{\bm{\mu}}}

\def\vtheta{{\bm{\theta}}}

\def\vphi{{\bm{\phi}}}
\def\vsigma{{\bm{\sigma}}}

\def\vphi{{\bm{\phi}}}

\def\mI{{\bm{I}}}

\DeclareMathAlphabet{\mathsfit}{\encodingdefault}{\sfdefault}{m}{sl}
\SetMathAlphabet{\mathsfit}{bold}{\encodingdefault}{\sfdefault}{bx}{n}

\newcommand{\diag}{\mathrm{diag}}
\newcommand{\E}{\mathbb{E}}

\icmltitlerunning{Discriminative Regularization for Generative Models}

\begin{document} 

\twocolumn[
\icmltitle{Discriminative Regularization for Generative Models}

\icmlauthor{Alex Lamb, Vincent Dumoulin and Aaron Courville}{first.last@umontreal.ca}
\icmladdress{Montreal Institute for Learning Algorithms,
			Universit\'{e} de Montr\'{e}al}

\icmlkeywords{Deep Learning, Machine Learning, ICML}
\vskip 0.3in
]

\begin{abstract}
We explore the question of whether the representations learned by classifiers can be used to enhance the quality of generative models. Our conjecture is that labels correspond to characteristics of natural data which are most salient to humans: identity in faces, objects in images, and utterances in speech.  We propose to take advantage of this by using the representations from discriminative classifiers to augment the objective function corresponding to a generative model. In particular we enhance the objective function of the variational autoencoder, a popular generative model, with a discriminative regularization term. We show that enhancing the objective function in this way leads to samples that are clearer and have higher visual quality than the samples from the standard variational autoencoders.  


\end{abstract}

\section{Introduction}

Discriminative neural network models have had a tremendous impact in many traditional application areas of machine learning such as object recognition and detection in images \cite{krizhevsky2012imagenet,simonyan2014vgg}, speech recognition \cite{hinton2012acoustic} and a host of other application domains \cite{schmidhuber2014deepoverview}. While progress in the longstanding problem of learning generative models capable of producing novel and compelling examples of natural data has not quite kept pace with the advances in discriminative modeling, there have been a number of important developments. 

Within the context of generative models that support tractable approximate inference, the variational autoencoder (VAE) \cite{kingma2013vae} has emerged as a popular framework. The VAE leverages deep neural networks both for the generative model (mapping from a set of latent random variables to a conditional distribution over the observed data) and for an approximate inference model (mapping from the observed data to a conditional distribution over the latent random variables).

Images generated from the VAE (and most other generative frameworks) diverge from natural images in two distinct ways: 
\begin{enumerate}
\item \emph{Missing high frequency information.} Compared to natural data, generated samples often lack detail and appear blurry. Generative models of natural data such as images are largely limited to the maximum likelihood setting where the data was modeled as Gaussian distributed (with diagonal covariance), given some setting of the latent variables. Under a Gaussian, the quality of reconstruction is essentially evaluated on the basis of a generalized $\mathrm{L}^2$ distance. As a measure of similarity between images, $\mathrm{L}^2$ distance does not closely match human perception. For instance, the same image translated by a few pixels could have relatively high $\mathrm{L}^2$ distance, yet humans may not even perceive the difference. 

\item \emph{Missing semantic information.} Human perception is goal driven: we perceive our environment so that we can interact with it in meaningful ways. This implies that semantic information is going to be particularly salient to the human perceptual system. The current state-of-the-art in generative models, even when they capture high frequency information, produce samples which often lack semantically-relevant details. Generative models of natural images often lack a clear sense of ``objectness''. It is not enough to capture the correct local statistics over the data. For example, generative models trained on faces often produce inconsistencies in gender and identity, which may be subtle in pixel space but immediately apparent to humans viewing the samples.  
\end{enumerate}

In this work we explore an alternative VAE training objective by augmenting the standard VAE lower bound on the likelihood objective with additional discriminative terms that encourage the model's reconstructions to be close to the data example in a representation space defined by the hidden layers of highly-discriminative, neural network-based classifiers. We refer to this strategy as discriminative regularization of generative models. 

In this effort we are heavily inspired by recently introduced texture synthesis method of \cite{gatys2015texture} as well as the DeepStyle model of \cite{gatys2015style}. These works showed that 
surprisingly detailed and semantically-rich information regarding natural images is preserved in the hidden-layer representations of ImageNet-trained object recognition networks such as VGG \cite{simonyan2014vgg}. Our goal is to incorporate this insight into the VAE framework and to render the synthetic data perceptually closer to the real data. 

In this paper we confine our discussion to learning generative models of images; however, the approach we propose here is readily applicable to other domains. We show how to learn discriminatively regularized generative models for three benchmark datasets: Street View House Numbers (SVHN) \cite{netzer2011reading}, the CIFAR-10 object recognition dataset \citep{krizhevsky2009learning}, and the CelebA facial attribute recognition dataset \cite{liu2015faceattributes}. In each case, the classifier we consider is a convolutional neural network (CNN).

\begin{figure*}[ht]
\centering
\input{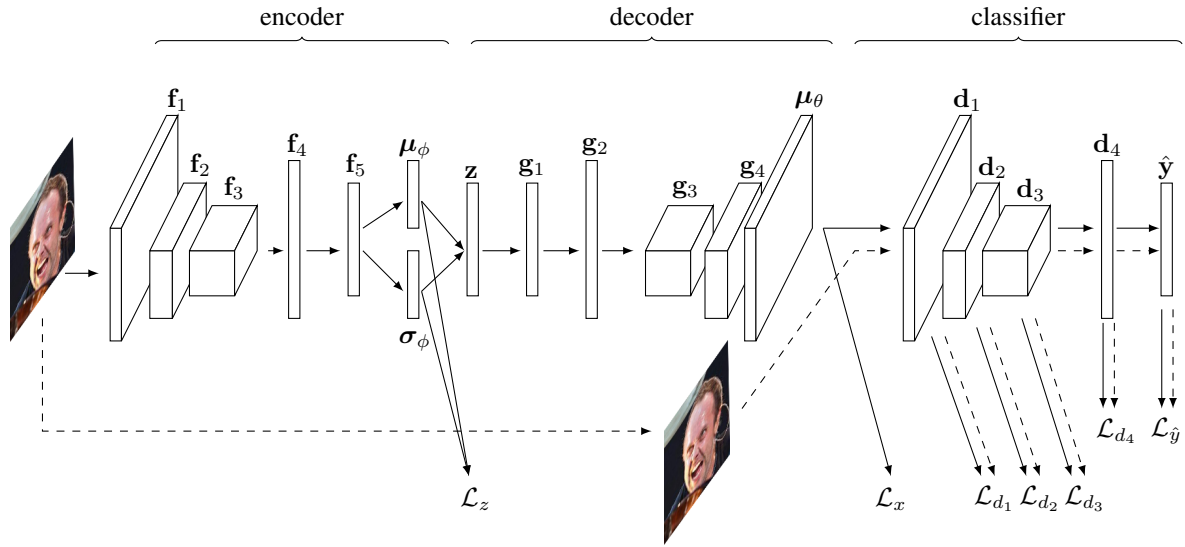}
\caption{\label{fig:model_pipe}The discriminative regularization model. Layers $\vf_1$, $\vf_2$, $\vf_3$, $\vd_1$, $\vd_2$ and $\vd_3$ represent convolutional layers, whereas layers $\vg_3$, $\vg_4$ and $\vmu_{\theta}$ represent fractionally strided convolutional layers.}
\end{figure*}

\section{VAEs as Generative models of images}
In this section we lay out the variational autoencoder (VAE) framework \cite{kingma2013vae,rezende2014stochastic} on which we build. The VAE is a neural network-based approach to latent variable modeling where the natural, richly-structured dependencies found in the data are disentangled into the relatively simple dependencies between a set of latent variables.  Formally, let $\rvx$ be a random real-valued vector representing the observed data and let $\rvz$ be a random real-valued vector representing the latent variables that reflect the principle directions of variation in the input data.

\subsection{The generative model}
 We specify the generative model over the pair $(\vx,\vz)$ as $p_{
\theta}(\vx,\vz) = p_{\theta}(\vx \mid \vz)p_{\theta}(\vz)$, where $p_{\theta}(\vz)$ is the prior distribution over the latent variables and $p_{\theta}(\vx \mid \vz)$ is the conditional likelihood of the data given the latents. $\theta$ represents the generative model parameters. As is typical in the VAE framework, we assume a standard Normal  (Gaussian) prior distribution over $\vz$: $p_{\theta}(\vz) = \mathcal{N}(\vz \mid \vzero,\mI)$. 

For real-valued data such as natural images, by far the most common conditional likelihood is the Gaussian distribution: $p(\vx \mid \vz) = \mathcal{N}(\vx \mid \vmu_{\theta}(\vz),\mathrm{diag}(\vsigma^2_{\theta}))$, where the mean $\vmu_x(\vz)$ is a nonlinear function of the latent variables specified by a neural network, which following autoencoder terminology, we refer to as the \emph{decoder network}, $f(\vx)$. In the natural image setting, $\vmu_{\theta}(\vz)$ is parameterized by a CNN (see \autoref{fig:model_pipe}) and $\vsigma^2_{\theta}$ is a vector of independent variance parameters over the pixels.

\subsection{The approximate inference model}
Given the generative model described above, inference is intractable, as is standard parameter learning paradigms such as maximizing the likelihood of the data. 
The VAE resolves these issues by introducing a learned
approximate posterior distribution $q_\phi(\mathbf{z} \mid \mathbf{x})$,
 specified by another neural network known as the \emph{encoder network}, $g(\vz)$ and parametrized by $\phi$. 

Introducing the approximate posterior $q_\phi(\mathbf{z} \mid \mathbf{x})$ allows us to decompose the marginal log-likelihood of the data under the generative model in terms of the variational free energy and the Kullback-Leibler divergence between the approximate and true posteriors:
\begin{equation}
\log p_{\vtheta}(\vx) = \mathcal{L}(\theta,\phi;\vx) + D_{\mathrm{KL}}\left(q_{\vphi}(\vz \mid \vx)\|p_{\vtheta}(\vz \mid \vx)\right) 
\label{eq:ELBO}
\end{equation}
where the Kullback-Leibler divergence is given by
\begin{equation}
D_{\mathrm{KL}}\left(q_{\vphi}(\vz \mid \vx) \| p_{\vtheta}(\vz \mid \vx)\right) = \E_{q_{\phi}(\vz \mid \vx)} \left[ \log \frac{q_{\vphi}(\vz \mid \vx)}{p_{\vtheta}(\vz \mid \vx)} \right] \notag
\end{equation}
and the variational free energy is given by
\begin{equation} 
\mathcal{L}(\theta,\phi;\vx)= \E_{q_{\vphi}(\vz \mid \vx)}\left[\log \frac{p_{\vtheta}(\vx,\vz)}{q_{\vphi}(\vz \mid \vx)} \right]. \notag
\end{equation}
Since $D_{\mathrm{KL}}\left(q_{\vphi}(\vz \mid \vx) \| p_{\vtheta}(\vz \mid \vx)\right)$ measures the divergence between $q_{\vphi}(\vz \mid \vx)$ and $p_{\vtheta}(\vz \mid \vx)$, it is guaranteed to be non-negative. As a consequence, the variational free energy $\mathcal{L}(\theta,\phi;\vx)$ is always a lower bound on the likelihood. As such it is sometimes called the variational lower bound or the evidence lower bound (ELBO). 

In the VAE framework, $\mathcal{L}(\theta,\phi;\vx)$ is often rearranged into two terms: 
\begin{equation} 
\mathcal{L}(\theta,\phi;\vx) =  \mathcal{L}_{\vz}(\theta,\phi;\vx) + \mathcal{L}_{\vx}(\theta,\phi;\vx)
\end{equation}
where
\begin{align}
\mathcal{L}_{\vz}(\theta,\phi;\vx) &= - D_{\mathrm{KL}}\left(q_{\vphi}(\vz \mid \vx) \| p_{\vtheta}(\vz)\right) \notag \\
\mathcal{L}_{\vx}(\theta,\phi;\vx) &= \E_{q_{\vphi}(\vz \mid \vx)}\left[ \log p_{\vtheta}(\vx \mid \vz) \right] \notag 
\end{align}
$\mathcal{L}_{\vx}$ can be interpreted as the (negative) expected reconstruction error of $\vx$ under the conditional likelihood with respect to $q_{\phi}(\vz \mid \vx)$.
Maximizing this lower bound strikes a balance between minimizing reconstruction error and minimizing the KL divergence between the approximate posterior $q_{\phi}(\vz \mid \vx)$ and the prior $p_{\theta}(\vz)$.

\subsection{Reparametrization Trick}

The power of the VAE approach can be credited to how the model is trained.
With real-valued $\vz$, we can exploit a \emph{reparametrization trick}
\mbox{\cite{kingma2013vae,bengio2013estimating}} to
propagate the gradient from the decoder network to the encoder network. Instead of sampling directly from $q_\phi(\vz
\mid \vx)$, $\vz$ is computed as a deterministic function of
$\vx$ and some noise term $\epsilon \sim \mathcal{N}(\vzero,\mI)$ such that $\mathbf{z}$ has the desired distribution. For instance, if
\begin{equation}
    q_\phi(\vz \mid \vx) = \mathcal{N}(\vz \mid \mu_\phi(\vx),
                                  \diag(\vsigma^2_\phi(\vx))),
\end{equation}
then we would express $\vz$ as
\begin{equation}
    \vz = \mu_{\phi}(\vx) + \vsigma_{\phi}(\vx) \odot \epsilon,
    \quad \epsilon \sim \mathcal{N}(0, I) \notag
\end{equation}
to produce values with the desired distribution while permitting gradients to propagate through both $\mu_\phi(\vx)$ and $\vsigma^2_{\phi}(\vx)$. 

\subsection{The problem with the Independent Gaussian Assumption}

The derivation of the variational autoencoder allows for different choices for the reconstruction model $p_{\theta}(\vx \mid \vz)$. However, as previously mentioned the majority of applications on real-valued data use a multivariate Gaussian with diagonal covariance matrix as the conditional likelihood of the data given the latent variables \cite{gregor2015draw,mansimov2015captions}.  Maximizing the conditional likelihood of this distribution corresponds to minimizing an elementwise $L^2$  reconstruction penalty.  One major weakness with this approach is that elementwise distance metrics are a poor fit for human notions of similarity.  For example, shifting an image by only a few pixels will cause it to look very different under elementwise distance metrics but will not change its semantic properties or how it is perceived by humans \cite{theis2015evaluation}.  

In addition to the issues surrounding elementwise independence, there is nothing in a Gaussian conditional likelihood that will cause the model to render semantically-salient perceptual features of the data to be captured by the model.

\section{Discriminative Regularization}

In this section we describe our modification to the VAE lower bound training objective. Our goal is to modify the VAE training objective to 
render generated images perceptually closer to natural images. As previously discussed, generated images from the VAE (or other generative frameworks) often diverge from natural images in two distinct directions: (1) by being excessively blurry and (2) by lacking semantically meaningful cues such as depictions of well-defined objects. We conjecture that both of these issues can be ameliorated by encouraging the generator to render reconstructions that match the original data example in a representation space defined by the hidden layers of a classifier trained on a discrimination task relevant to the input data.

Let $\vd_1(\vx),\vd_2(\vx),\dots,\vd_L(\vx)$ represent the $L$ hidden layer representations of a \emph{pre-trained} classifier. The classifier could be trained on a task specifically relevant to the data we wish to model. For example, in learning to generate images of faces we may wish to leverage a classifier trained to either identify individuals \cite{Huang2007LFW} or trained to recognize certain facial characteristics \cite{liu2015faceattributes}. On the other hand, we could also follow the example of \cite{gatys2015texture} and use one of the  high performing ImageNet trained models such as VGG \cite{simonyan2014vgg} as a general purpose classifier for natural images. 

In the standard VAE variational lower bound objective, we include a term that aims to minimize the reconstruction error in the space of the observed data. To this we add additional terms aimed at minimizing the reconstruction error in the space defined by the hidden layer representations, $\vd_1,\dots,\vd_L$, of the classifier. 
\begin{align}
\mathcal{L}_{disc}(\theta,\phi;\vx) = &-\mathcal{L}_{\vz}(\theta,\phi;\vx) + \mathcal{L}_{\vx}(\theta,\phi;\vx) \notag \\
&+ \sum_{l=1}^{L} \mathcal{L}_{\vd_l}(\theta,\phi;\vx),
\end{align}
where
\begin{equation}
\mathcal{L}_{\vd_l}(\theta,\phi;\vx) = \E_{q_{\phi}(\vz \mid \vx)} \log p_{\theta}(\vd_{l}(\vx) \mid \vz).
\end{equation}
We take the conditional likelihood of each $\vd_{l}(\vx) \mid \vz$ to be Gaussian with its mean $\vmu_{\vd_{l}}(\vz)$ defined by forward propagating the conditional mean $\vmu_{x}(\vz)$ through the layers of the classifier from $\vd_1$ to $\vd_l$:
\begin{align}
\vd_{1}(\vx) \mid \vz &\sim \mathcal{N}((\vd_1 \circ \vmu_{\theta})(\vz),\diag(\vsigma^2_{d,1})), \notag \\
\vd_{2}(\vx) \mid \vz &\sim \mathcal{N}((\vd_2 \circ \vd_1 \circ \vmu_{\theta})(\vz)  ,\diag(\vsigma^2_{d,2})), \notag \\
\dots \notag \\
\vd_{L}(\vx) \mid \vz &\sim \mathcal{N}((\vd_L \circ \dots \circ \vd_2 \circ \vd_1 \circ \vmu_{\theta})(\vz) ,\diag(\vsigma^2_{d,L})). \notag
\end{align}

The discriminative regularization approach can be considered a kind of multitask regularization of the standard VAE, where in addition to the standard VAE objective, we include the additional tasks of predicting each of the hidden layer representations of a classifier.

We can understand the impact that these additional terms would have on the VAE parameters by considering matching in the different layers of the classifier. Since the classifiers we will consider will all be convolutional neural networks, the different layers will tend to have different characteristics, especially with respect to spatial translations. 
Matching the lower layer representations is going to encourage visual features such as edges to be well-defined and in the right location. The upper layers of a convolutional neural network classifier have been shown to be both highly invariant to spatial transformations (particularly translation), while simultaneously showing high specificity to semantically-relevant stimuli. 
Matching in the upper layers will likely de-emphasize exact spatial alignment, but will pressure semantic elements apparent in the example, such as the identity of objects, to be well matched between the data example $\vx$ and the mean of the conditional likelihood $\vmu_{x}$.

It is important to assess the impact that the addition of our discriminative regularization terms have on the VAE. By adding the discriminative regularization terms we are no longer directly optimizing the variational lower bound. 

Furthermore, since we are backpropagating the gradient of the combined objective $\mathcal{L}_{disc}$ through the decoder network and into the encoder network (the network responsible for approximating the posterior distribution), we are no longer directly optimizing the encoder network to minimize $\mathrm{KL}\left(q(\vz \mid \vx),p(\vz \mid \vx)\right)$. Doing so implies that we risk deteriorating our approximate posterior in favor of improving the example reconstructions (w.r.t the combined objective). One consequence could be an overall deterioration of the generated sample quality as the marginal $q(\vz) = \int q(\vz \mid \vx)q(\vx)\ d\vx$ diverges from the prior $p(\vz)$. 

In our experiments, we did not observe any negative impact in sample quality, however if such an issue did arise, we could simply have elected not to propagate the the gradient contribution due to our discriminative regularization through the encoder network and thus preserve direct minimization of $\mathrm{KL}\left(q(\vz \mid \vx),p(\vz \mid \vx)\right)$ w.r.t. the parameters of the encoder network.

\begin{figure*}[ht]
\centering
\subfigure[Samples without discriminative regularization]{\includegraphics[scale=0.45]{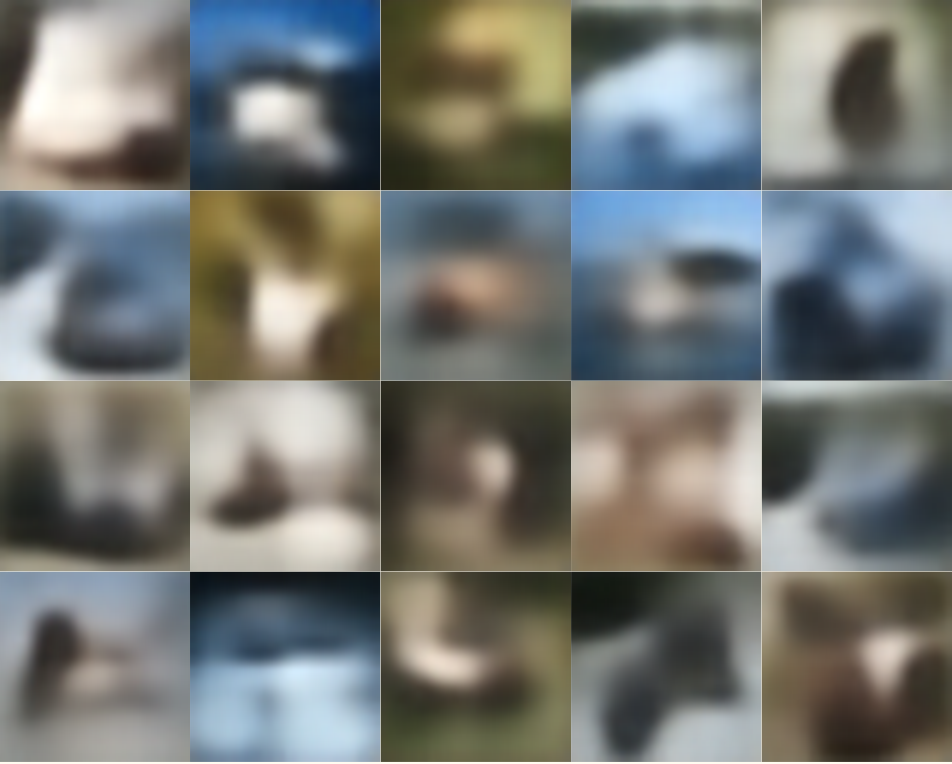}}\quad
\subfigure[Samples with discriminative regularization]{\includegraphics[scale=0.45]{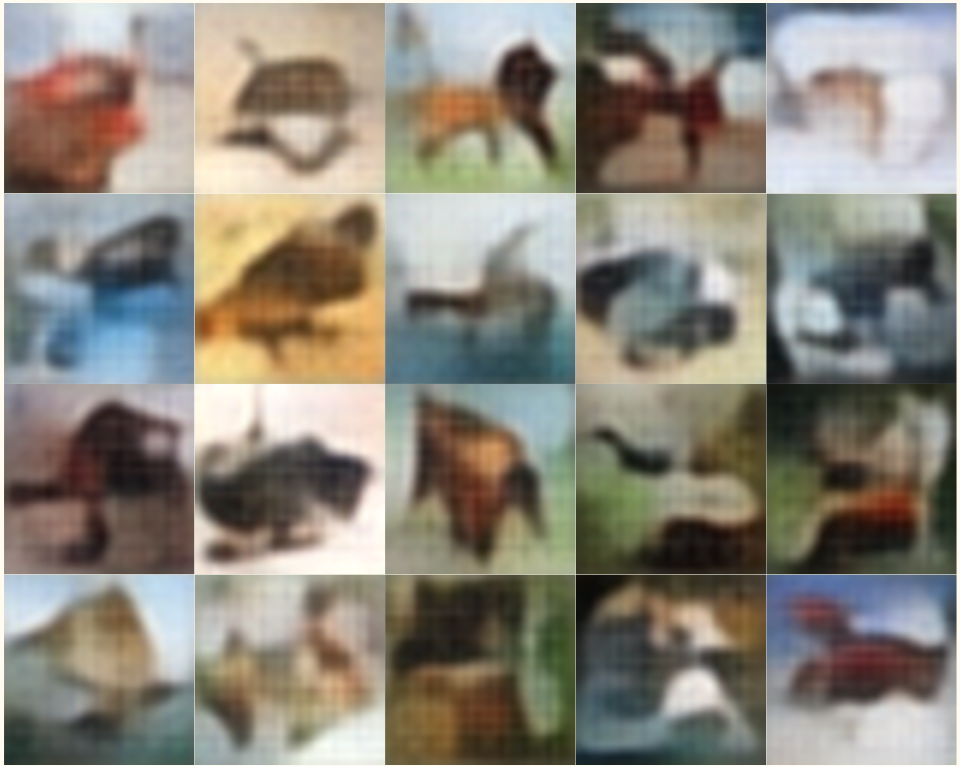}}\quad
\caption{\label{fig:cifar10_samples} CIFAR samples generated from variational autoencoders trained with and without the discriminative regularization.  The architecture and the hyperparameters (except those directly related to discriminative regularization) are the same for both models.  Our baseline VAE samples are similar in visual fidelity to other results in the literature \cite{mansimov2015captions}.  Discriminative regularization often does a good job of producing coherent objects, but the textures are usually muddled and the samples lack local detail}
\end{figure*}

\begin{figure*}[ht]
\centering
\includegraphics[scale=0.45]{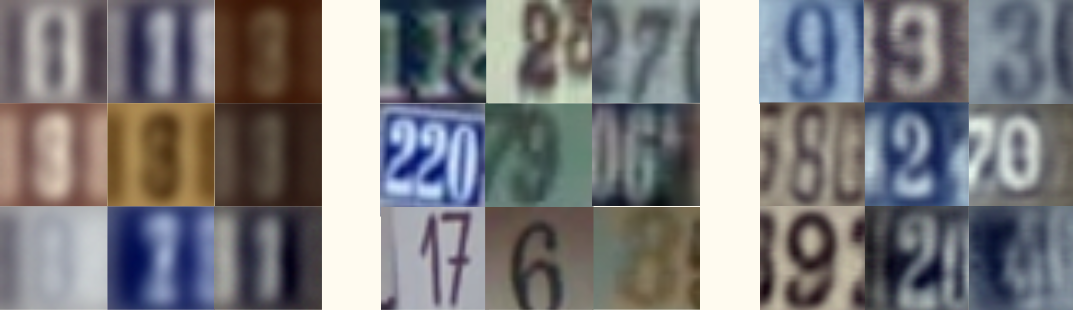}\quad
\caption{\label{fig:svhn_samples} SVHN samples with the standard variational autoencoders (left), real images (center), and samples using discriminative regularization (right).  The discriminative regularizer improves the clarity and visual fidelity of the samples.  SVHN is the only dataset where we did not observe unnatural patterning when using discriminative regularization.}
\end{figure*}

\section{Related Work}

Recent work has used the structural similarity metric \cite{wang2004ssim} as an auxiliary loss function for training variational autoencoders \cite{ridgeway2015perceptual}.  They showed that using this metric instead of pixel-wise square loss dramatically improved human ratings of the generated images.  Our approach differs from theirs in a few ways.  First, we use the representations from a discriminatively trained classifier to augment our objective function, whereas they use a hand-crafted measure for image similarity.  Second, discriminative regularization describes both local and global properties of the image (the local properties coming from lower layers and the global properties coming from higher layers), whereas their method only compares the true image and the reconstructed image around local 11x11 patches centered at each pixel.  An interesting area for future work would be to study which method does a better job at improving the generation of local data, or if results can be improved by using both methods simultaneously.  

Recently there has been a focus on alternative measures to be used during the training of generative models. Probably the most established of these is the generative adversarial networks (GANs) that leverage discriminative machinery and apply it to a two player game scenario between a generator and a discriminator \cite{goodfellow2014gan}. While the discriminator is trained to distinguish between true training samples and those generated from the generator, the generator is trained to try to fool the discriminator. While this joint optimization of the generator and discriminator is prone to instabilities, the end result are often generated images that capture realistic local texture. Recent applications of the GAN formalism have show very impressive results \cite{denton2015lapgan,radford2015dcgan}. 

Of all the proposed GAN-based methods, the one that most closely resembles the approach we propose here is the discriminative VAE \cite{larsen2015autoencoding}. In this work, the authors integrate the VAE within a GAN framework, in part, by maximizing a lower bound on a representation of the image defined by a given hidden layer of the GAN discriminator network. The authors show that their integration of the GAN and the VAE leads to impressive samples.  

While generative adversarial networks have been a driving force in the relatively rapid improvement in the quality of image generation models, there are ways in which VAEs are preferable. GAN models do not optimize likelihood and are not trained directly for coverage of the training set, i.e. they use their capacity to convincingly mimic natural images.  On the other hand the VAE more explicitly encourages coverage by maximizing a lower bound on the log likelihood.  Another disadvantage of GANs is that in their original formulation there is no clear way to perform inference in the model, i.e. to recover the posterior distribution $p(z \mid x)$. However, there has been a few very recent efforts  that are working to address this shortcoming of the GAN framework \cite{makhzani2015adversarial,larsen2015autoencoding}.  

\begin{figure*}[ht]
\centering
\subfigure[Samples without discriminative regularization]{\includegraphics[scale=0.45]{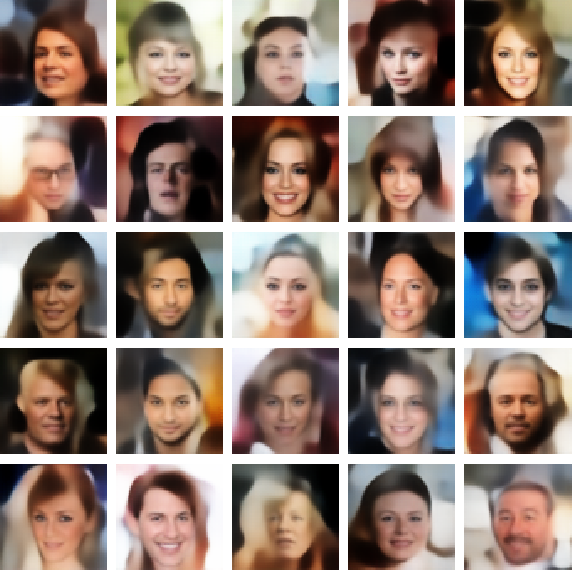}}\quad
\subfigure[Samples with discriminative regularization]{\includegraphics[scale=0.45]{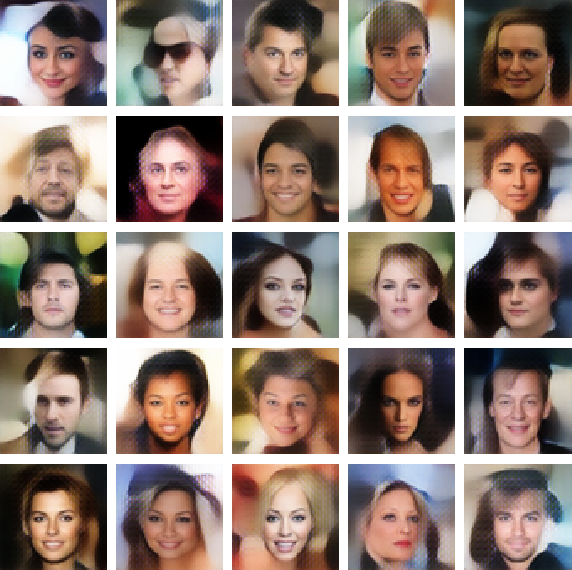}}\quad
\caption{\label{fig:celeba_samples} Face samples generated with and without discriminative regularization. On balance, details of the face are better captured and more varied in the samples generated with discriminative regularization.}
\end{figure*}

\begin{figure*}[ht]
\centering
\includegraphics[width=0.8\textwidth]{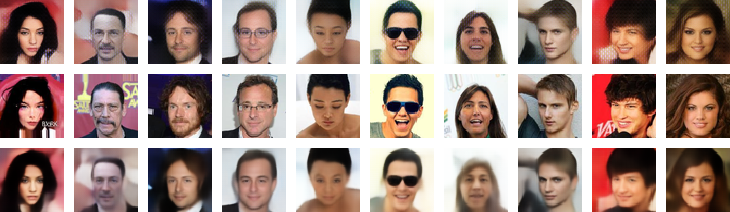}
\caption{\label{fig:celeba_reconstructions}Face reconstructions with (top row) and without (bottom row) discriminative regularization.  The face images used for the reconstructions (middle row) are from the held-out validation set and were not seen by the model during training.  The architecture and the hyperparameters (except those directly related to discriminative regularization) are the same for both models.  Discriminative regularization greatly enhances the model's ability to preserve identity, ethnicity, gender, and expressions.  Note that the model does not improve the visual quality of the image background, which likely reflects the fact that the classifier's labels all describe facial attributes.  Additional reconstructions can be seen in the appendix.  }
\end{figure*}

\section{Experiments}

We evaluated the impact of the discriminative regularization on three datasets: CelebA \cite{liu2015deep}, Street View House Numbers \cite{netzer2011reading}, and CIFAR-10 \cite{krizhevsky2009learning}. The SVHN and CIFAR-10 datasets were used as is, whereas the aligned and cropped version of the CelebA dataset was scaled from $218 \times 178$ pixels to $78 \times 64$ pixels and center cropped at $64 \times 64$ pixels.  For SVHN and CIFAR-10 we used the pre-trained VGG-19 model as the network for our discriminative regularization \cite{simonyan2014vgg} and for celebA we trained our own classifier to predict all of the labels.  

\begin{table}[h]
\centering
\small
\begin{tabular}{| l | c | c |}\hline
    5 Shadow & Arch. Eyebrows & Attractive \\
    Bags under eyes & Bald & Blurry \\
    Bangs & Big Lips & Brown Hair \\
    Big Nose & Black Hair & Bushy Eyebrows \\
    Blond Hair & Goatee & Gray Hair \\
    Eyeglasses & Double Chin & Heavy Makeup \\
    Heavy Cheekbones & Gender & Mouth Open \\
    Mustache & Narrow Eyes & No Beard \\
    Oval Face & Pale Skin & Pointy Nose \\
    Recced. Hairline & Rosey Cheeks & Sideburns \\
    Smiling & Straight Hair & Wavy Hair \\
    Earrings & Wearing Hat & Lipstick \\
    Necklace & Necktie & Young \\
    \hline
\end{tabular}
\caption{\label{tab:celeblabels}A list of the binary targets that we predict with our celebA classifier.}
\end{table}

All VAE models, regularized or not, as well as the CelebA classifier were trained using Adam and batch normalization.  Our architecture closely follows \cite{radford2015dcgan}, with convolutional layers in the encoder and fractionally-strided convolutions in the decoder.  In each convolutional layer in the encoder we double the number of filters present in the previous layer and use a convolutional stride of 2.  In each convolutional layer in the decoder we use a fractional stride of 2 and halve the number of filters on each layer.  

Evaluating generative models quantitatively is a challenging task \cite{2015arXiv151101844T}.  One common evaluation metric is the likelihood of held-out samples.  However, the usefulness of this metric is limited.  If we compare the log-likelihood using the independent Gaussian in the pixel space, then we suffer from the limitations of pixel-wise distance metrics for comparing images.  On the other hand, if we compare using the log-likelihood over the hidden states of the discriminative classifier, then we bias our evaluation criteria towards the criteria that we trained on.  

\begin{figure*}[ht]
\centering
\includegraphics[width=0.8\textwidth]{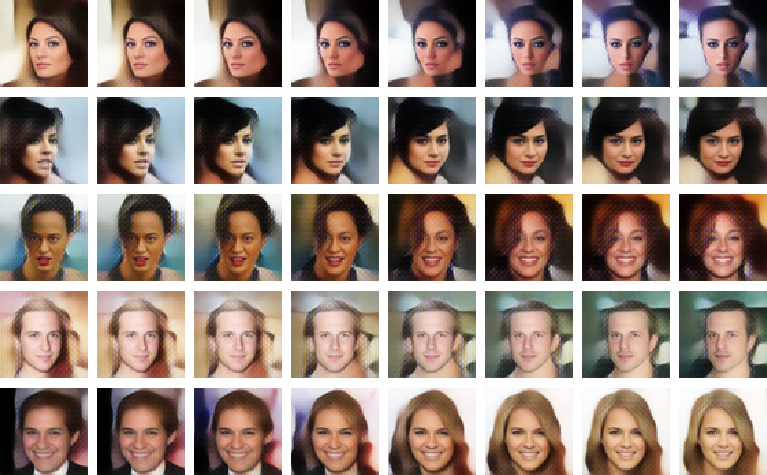}
\caption{\label{fig:celeba_interpolations} Latent space interpolations with  discriminative regularization. On each row, the first and last image correspond to reconstructions of randomly selected examples.}
\end{figure*}

\subsection{Samples}

Trained models were sampled from by sampling $\vz \sim p_{\theta}(\vz)$ and computing $\E_{q_{\phi}(\vz \mid \vx)}[p(\vx \mid \vz)]$ (in our case $\vmu_{\theta}(\vz)$), which is standard practice in generative modeling work.

Using discriminative regularization during training has a noticeable impact on the quality of CIFAR-10 samples (\autoref{fig:cifar10_samples}). In addition to being sharper, the samples also exhibit good global statistics, i.e. they look like objects.  We observe a similar improvement in the quality of our SVHN samples.  

Faces in CelebA samples (\autoref{fig:celeba_samples}) look more ``in focus'' when discriminative regularization is used during training.

\subsection{Quantitative Results}

In \autoref{tab:nll}, we show NLL approximations of models trained on the CelebA dataset with and without discriminative regularization. We report per-unit averages.  The approximation was obtained via importance sampling using 100 samples per data point. 

\begin{table}[h]
\centering
\begin{tabular}{l | c | c}
    Split & Without disc. reg. & With disc. reg. \\
    \hline
    & & \\
    Training & -1.2092 & -1.1147 \\
    Validation & -1.1779 & -1.0804 \\
    Test & -1.1835 & -1.0866 \\
\end{tabular}
\caption{\label{tab:nll}NLL approximations for models trained on the CelebA dataset with and without discriminative regularization.  We note that the discriminative regularizer makes the likelihood over the raw pixel space worse even though the visual quality of the samples is improved.}
\end{table}

\subsection{Reconstructions}

Reconstructions were obtained by sampling $\vz \sim q_{\phi}(\vz \mid \vx)$ and computing $\E_{q_{\phi}(\vz \mid \vx)}\left[p(\vx \mid \vz)\right]$ (in our case $\vmu_{\theta}(\vz)$), which is also standard practice in generative modeling work.

Using discriminative regularization during training leads to improved reconstructions (\autoref{fig:celeba_reconstructions}). In addition to producing sharper reconstructions, this approach helps maintaining the identity better. This is especially noticeable in the eyes region: VAE reconstructions tend to produce stereotypical eyes, whereas our approach better captures the overall eye shape.

\subsection{Interpolations in the Latent Space}

To evaluate the quality of the learned latent representation, we visualize the result of linearly interpolating between latent configurations. We choose pairs of images whose latent representation we obtain by computing $\vmu_{\phi}(\vx)$. We then compute intermediary latent representations $\vz$ by linearly interpolating between the latent representation pairs, and we display the corresponding $\vmu_{\theta}(\vz)$.

The resulting trajectory in pixel space (\autoref{fig:celeba_interpolations}) exhibits smooth and realistic transitions between face pose and orientation, hair color and gender.

\subsection{Explaining Visual Artifacts}

In the samples generated from a model trained with discriminative regularization, we sometimes see unnatural patterns or texturing.  In the faces samples, we mostly observe these patterns in the background.  In the CIFAR dataset, they occur to some extent in nearly all samples.  These patterns are not seen in samples from the standard variational autoencoders. 

One explanation for the visual artifacts is that the variational autoencoder with discriminative regularization produces unnaturally blurred activations in the classifier's convolutional layers in the same way that the standard variational autoencoder outputs unnaturally blurred images.

To support this hypothesis, we visualize what happens when a convolutional autoencoder explicitly tries to generate a reconstruction which produces a blurred representation in the classifier. To do so, we train a convolutional autoencoder on a batch of 100 examples. The examples are reconstructed as usual, but we propagate both the input and the reconstruction through the first two layers of the classifier. The propagated input is then blurred by adding gaussian blur (applied separately to each filter), and the cost is computed as the squared error between the propagated reconstruction and the blurred propagated input.

\begin{figure}
\centering
\includegraphics[width=0.8\linewidth]{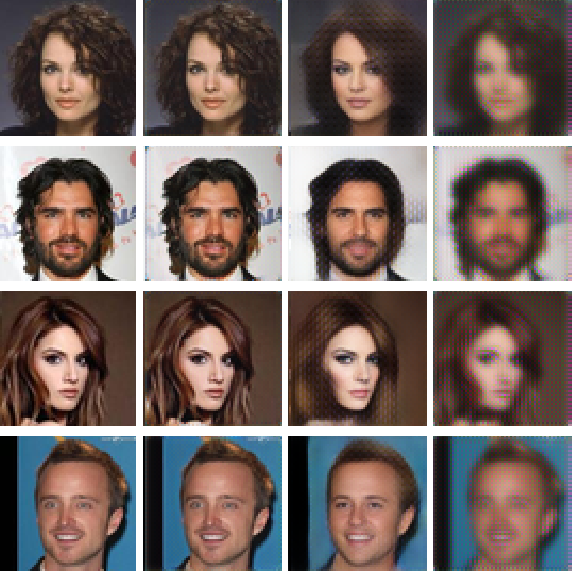}
\caption{\label{fig:patterning}From left to right: input examples, convolutional (non-variational) autoencoder reconstructions (no blurring applied to the classifier's hidden representations), model reconstructions (trained with discriminative regularization),  convolutional autoencoder reconstructions (blurring applied to the classifier's hidden representations).}
\end{figure}

\autoref{fig:patterning} provides a visual summary of the experiment. We see that when no blurring is applied to the hidden representation, the autoencoder does a perfect job of matching the hidden representations (middle left column), which is indicated by an excellent reconstruction at the input level. When blurring is applied, we see that the resulting reconstructions (right column) exhibit visual patterns resembling those of our model's reconstructions (middle right column).

\section{Conclusion}

A common view in cognitive science is that generative modeling will play a central role in the development of artificial intelligence by enabling feature learning where labeled data and reward signals are sparse.  In this view generative models serve to assist other models by learning representations and discovering causal factors from the nearly unlimited supply of unlabeled data.  Our paper shows that this interaction ought to be a two-way street, in which supervised learning contributes to generative modeling by determining which attributes of the data are worth learning to represent. We have demonstrated that discriminative information can be used to regularize generative models to improve the perceptual quality of their samples. 





\bibliography{discgen}
\bibliographystyle{icml2016}

\end{document}